\documentclass[10pt,twocolumn,letterpaper]{article}

\usepackage[pagenumbers]{cvpr} %

\usepackage{graphicx}
\usepackage{amsmath}
\usepackage{amssymb}
\usepackage{booktabs}
\usepackage[accsupp]{axessibility}
\usepackage[pagebackref,breaklinks,colorlinks]{hyperref}

\usepackage[capitalize]{cleveref}
\crefname{section}{Sec.}{Secs.}
\Crefname{section}{Section}{Sections}
\Crefname{table}{Table}{Tables}
\crefname{table}{Tab.}{Tabs.}

\def\cvprPaperID{8123} %
\def\confName{CVPR}
\def\confYear{2022}

\usepackage{overpic}
\usepackage{enumitem} %
\usepackage{overpic} %
\usepackage{color}

\definecolor{turquoise}{cmyk}{0.65,0,0.1,0.3}
\definecolor{purple}{rgb}{0.65,0,0.65}
\definecolor{dark_green}{rgb}{0, 0.5, 0}
\definecolor{orange}{rgb}{0.8, 0.6, 0.2}
\definecolor{red}{rgb}{0.8, 0.2, 0.2}
\definecolor{darkred}{rgb}{0.6, 0.1, 0.05}
\definecolor{blueish}{rgb}{0.0, 0.3, .6}
\definecolor{light_gray}{rgb}{0.7, 0.7, .7}
\definecolor{pink}{rgb}{1, 0, 1}
\definecolor{greyblue}{rgb}{0.25, 0.25, 1}

\DeclareMathOperator*{\argmin}{arg\,min}

\usepackage{blindtext}

\renewcommand{\paragraph}[1]{\vspace{1em}\noindent\textbf{#1}.}

\begin{document}
\title{Optimal Correction Cost for Object Detection Evaluation}

\author{Mayu Otani\\
CyberAgent, Inc.\\
\and
Riku Togashi\\
CyberAgent, Inc.\\
\and
Yuta Nakashima\\
Osaka University\\
\and
Esa Rahtu\\
Tampere University\\
\and
Janne Heikkil\"{a}\\
University of Oulu\\
\and
Shin'ichi Satoh \\
CyberAgent, Inc. \\
}

\maketitle
\begin{abstract}
Mean Average Precision (mAP) is the primary evaluation measure for object detection.
Although object detection has a broad range of applications, mAP evaluates detectors in terms of the performance of ranked instance retrieval.
Such the assumption for the evaluation task does not suit some downstream tasks.
To alleviate the gap between downstream tasks and the evaluation scenario, we propose Optimal Correction Cost (OC-cost), which assesses detection accuracy at image level.
OC-cost computes the cost of correcting detections to ground truths as a measure of accuracy. The cost is obtained by solving an optimal transportation problem between the detections and the ground truths.
Unlike mAP, OC-cost is designed to penalize false positive and false negative detections properly, and every image in a dataset is treated equally.
Our experimental result validates that OC-cost has better agreement with human preference than a ranking-based measure, \ie, mAP for a single image.
We also show that detectors' rankings by OC-cost are more consistent on different data splits than mAP.
Our goal is not to replace mAP with OC-cost but provide an additional tool to evaluate detectors from another aspect.
To help future researchers and developers choose a target measure, we provide a series of experiments to clarify how mAP and OC-cost differ.
\end{abstract}
\section{Introduction}
\label{sec:intro}

Evaluation measure is an important factor determining the direction of the algorithm development. 
Most object detection benchmarks adopt the mean average precision (mAP) as their primary evaluation metric, and, therefore, great efforts are made to achieve higher mAP scores.
While much research relies on mAP, we may not be completely aware of the consequences of optimising mAP.

Mean average precision is a ranking measure used in the information retrieval community \cite{voorhees2002overview}. Originally, the VOC challenge adopted mAP and included it into the evaluation protocol for object detection \cite{Everingham2010}.
In this evaluation protocol, all detected instances are ranked in the order of their confidence scores. Then, the average precision (AP) for each object category is calculated from the precision/recall curve of the ranked instances. Mean average precision summarizes the individual APs by averaging them across all categories.
Evaluation using mAPs views object detection as a task to rank detected instances for each category.
However, the range of real-world applications of detection algorithms is broad, and the ranking measures may not always be the appropriate objective to be optimized.

\begin{figure}[t!]
\begin{center}
\includegraphics[width=\linewidth]{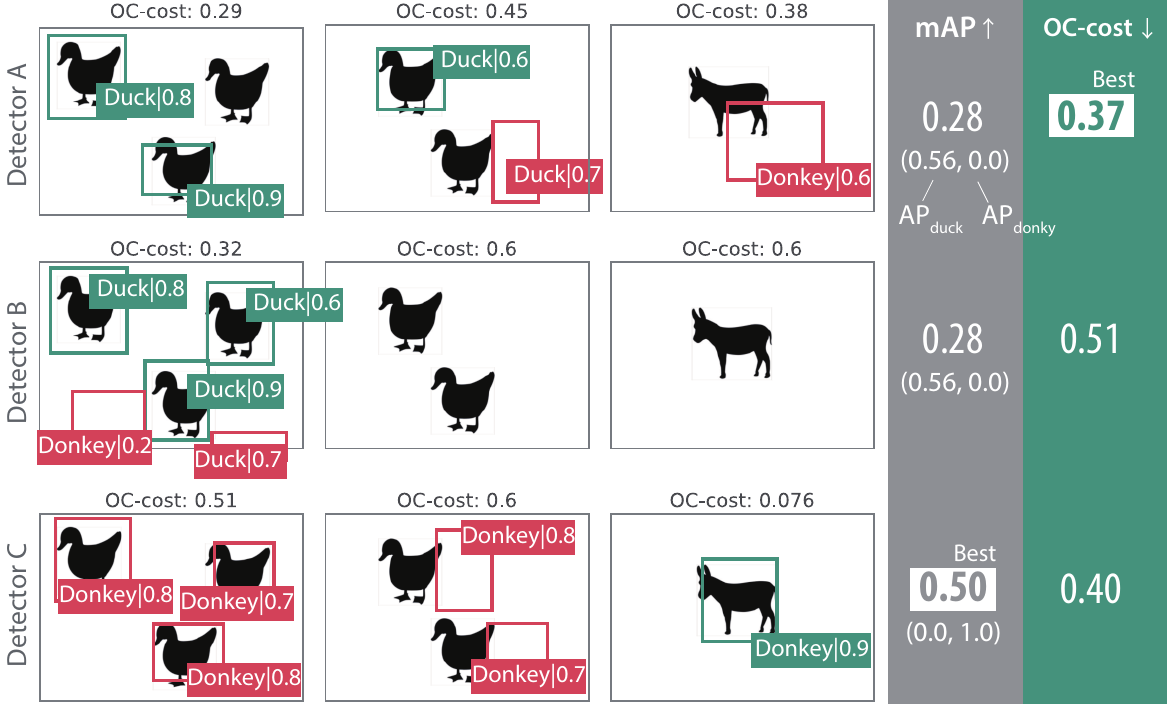}
\end{center}
\caption{
Toy example of three detectors and corresponding mAP and OC-cost.
Higher is better in mAP, and lower is better in OC-cost.
Top and middle: mAP does not treat each image equally. Detector A and B get the same mAP even though detector B does not produces any detections in the two images.
Bottom: mAP does not penalizes incorrect detections ranked lower than correct ones.
}
\label{fig:teaser}
\end{figure}
\Cref{fig:teaser} illustrates the characteristic behaviors of the mAP measure.
Suppose we have three detectors A, B, and C. 
Each detector attempts to find ducks and donkeys from a database consisting of three images.
Detectors A and B result in the identical mAP scores, although detector B completely ignores two of the three images.
This example demonstrates that mAP does not treat images equally, and a detector does not get penalized for not producing any detections in some images in the dataset.
Consistent operation on variety of images is a critical property in many real-world applications.
For example, in autonomous driving, neglecting the performance in rare scenes may lead to serious risks. 
However, mAP does not capture such local performance drop.
One remedy to mitigate this problem is to compute mAP for individual images by considering each image as a dataset, which has only one sample. 

There is another issue in ranking measure-based evaluation.
The example of detector C illustrates how mAP ignores some types of incorrect detections, which is non-intuitive.
Detector C is an extreme example that only detects donkeys.
Detector C produces many incorrect donkey detections, however, mAP does not penalize the incorrect detections as the detections are ranked lower than the correct one.
Although the mAP's behavior is reasonable for evaluation of a ranking problem such as content-based image retrieval, some applications need different type of evaluation.
For example, services like an on-demand visual recognition API, where various users independently upload their images, have to provide consistent performance for a wide range of images.
For such services, the detection accuracy in the image level is more important than the per-class ranking performance over the entire dataset.

There are also several problems in the mAP's implementation.
In mAP, each prediction is assigned to a ground truth to determine if the prediction is successful or not.
However, the assignment is obtained in a greedy fashion, which the obtained solution may not be optimal.
Non-optimal assignment may underrate detection results.
Furthermore, mAP commonly uses thresholding on the intersection over union (IoU) scores to determine the success or failure of each detection.
Prior research has shown that due to this thresholding, mAP does not reflect how well the predicted bounding box localizes the ground-truth instance \cite{LRP_kemal_2018}.
Lastly, in mAP, classification is more critical than the localization quality.
All detections are first grouped based on the predicted category, and misclassifications are considered as complete failures regardless of their localization quality.

In this paper, we propose a new evaluation measure called \emph{Optimal Correction Cost} (OC-cost) that aims to evaluate the detection accuracy at the image level.
Different from mAP that evaluates ranked instances detected in a whole dataset, OC-cost evaluates detection result for a single image, and the score is independent from other images.
Specifically, we evaluate the detection performance using the cost of correcting detections to the ground truths.
We expect that our evaluation measure better suits applications where image-level detection accuracy is critical.

To address the aforementioned problems, we formulate the computation of OC-cost as an optimal transportation problem.
For every detection and ground-truth pair, we define a unit correction cost that consists of a classification and localization cost.
Given the pair-wise correction costs, we find the globally optimal assignment that minimizes the total cost by solving an optimal transportation problem.
This approach, inspired by the previous work \cite{Ge_2021_CVPR}, avoids non-optimal assignments of detections to ground truths and IoU thresholding. The approach also allow users to balance classification and localization assessment.
We explore the potential of the optimal transportation cost as an alternative evaluation measure for detection tasks and re-define it as an evaluation measure.

Our contributions are summarized as follows:
\begin{itemize}
\item We develop an alternative evaluation measure for object detection tasks. Unlike mAP, which evaluates the performance of instance ranking, ours evaluates image-level detection accuracy. Image-level evaluation is suitable for some applications where a detector is expected to work consistently over various images.
\item We conduct a series of experiments to illustrate the behavior of OC-cost and how OC-cost differs from a ranking measure-based evaluation, \ie, mAP. We also demonstrate that our measure is useful for developing detectors by tuning post-processing parameters.
\end{itemize}
\section{Related Works}
\label{sec:related}

PASCAL VOC challenge introduced mAP for object detection evaluation \cite{Everingham2010}.
Given a ranked list of all detected instances, a precision/recall curve is computed.
mAP summarizes the precision/recall curve by computing the mean precision at equally spaced recall values.
Each instance is labeled as true positive if the instance's IoU is larger than a predefined threshold.
However, the binary judgment based on thresholding lacks the capability to describe the localization quality.
To alleviate the problem, a variant of mAP is proposed for COCO object detection task.
COCO-style mAP \cite{10.1007/978-3-319-10602-1_48} averages mAPs over multiple IoU thresholds ranging from 0.5 to 0.95.
\begin{figure*}[t!]
\begin{center}
\includegraphics[width=.9\linewidth]{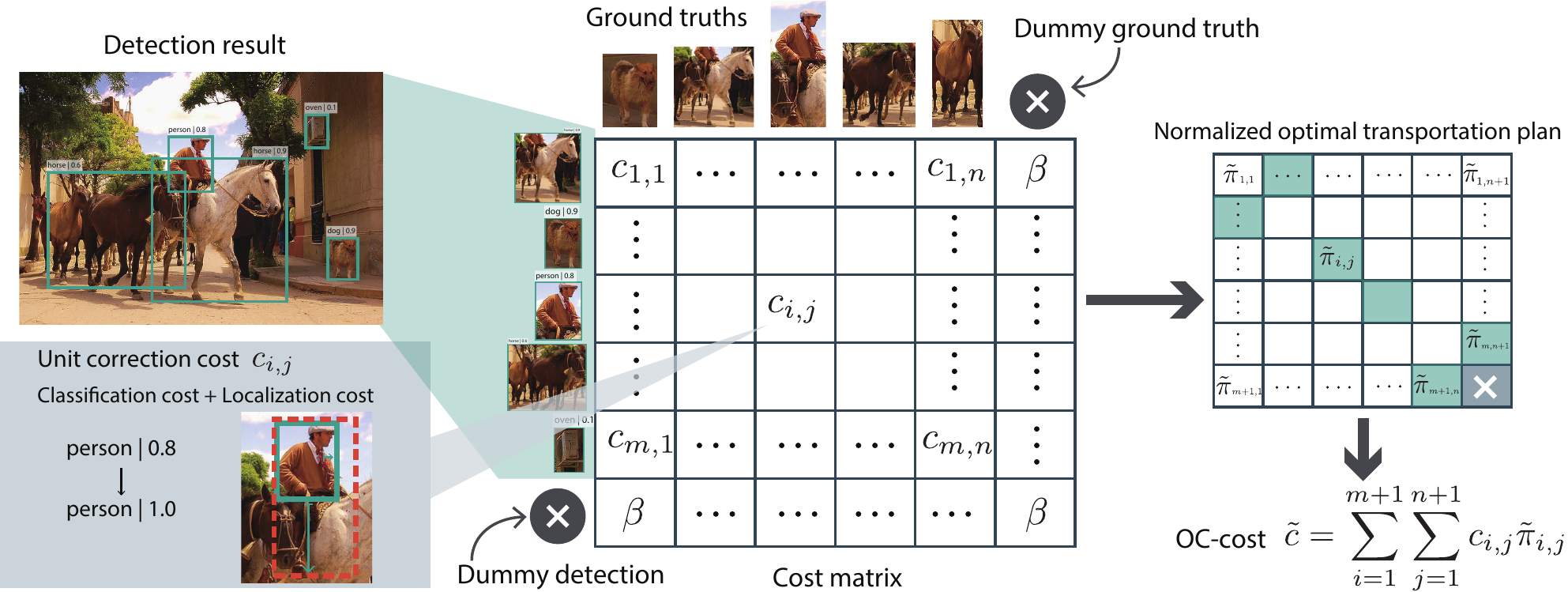}
\end{center}
\caption{
Overview of OC-cost. Given a set of detections and ground truths, we construct a cost matrix. Each element represents the cost of correction for a pair of a detection and a ground truth. We obtain optimal assignment of the detections to the ground truths by solving an optimal transportation problem. Based on the assignment, we aggregate the correction costs as the measure of detection performance.
}
\label{fig:overview}
\end{figure*}

Some previous works have proposed fixed measures to mitigate the limitations of mAP.
LRP \cite{LRP_kemal_2018} is designed to distinguish the characteristics of precision/recall curves.
LRP also introduces the average IoU of true positive instances in the measure to reflect localization quality.
Another research group pointed out that, due to a certain implementation of mAP, AP is not category independent \cite{Dave2021}.
Specifically, the limitation on detections per image can eliminate detections of rare categories with low confidence scores. As a result, mAP can be substantially degraded. 
They proposed a fixed evaluation protocol that limits the number of detections per category instead of limiting detections per image.
They also proposed a variant of mAP, which evaluates cross-category ranking.
Their mAP variant is employed by LVIS’21 \cite{Gupta_2019_CVPR}.
TIDE tries to reveal how a detector fails by computing mAP drop caused by certain types of errors, \eg, localization error and misclassification \cite{TIDE_derek_2012}.
Probabilistic object detection indicates another direction of evaluation \cite{Hall_2020_WACV}.
They proposed a new format of detection that requires to report uncertainty of detections.
Their measure evaluates if the detector can accurately estimate the uncertainty of detection.

Our measure is inspired by two prior works.
The first one is building detection evaluation \cite{OZDEMIR20101128}.
The evaluation measure matches predicted segments of buildings and references by solving a bipartite graph matching problem, then computes a shape-aware distance between matched segments.
The second work is a recently proposed learning method for object detection \cite{Ge_2021_CVPR}.
Their loss function is computed by solving an optimal transportation problem between predicted anchors and ground truths.
Their loss function aims to promote the proper assignment of anchors to supervision during training.
Since this loss function is not designed to evaluate the final detection results, it is not straightforward to use their formulation for evaluation.
Their formulation assumes one-to-many matching where each ground truth can be assigned to multiple anchors, and the loss values for different samples are not comparable.
To extend the idea to an image-level detection evaluation, we reformulate the optimal transportation problem.

\section{Optimal Correction Cost-based Measure}

For image-level evaluation, we consider costs to correct detections as a performance measure.
Figure \ref{fig:overview} illustrates the overview of our measure.
We assess the cost of correction for every detection and ground truth pair.
To compute the cost, we evaluate classification and localization errors.
We then construct a cost matrix consisting of the pairwise costs.
To find an optimal assignment of the detections to the ground truths, we solve an optimal transportation problem on the cost matrix \cite{10.1145/2070781.2024192}, then compute the OC-cost on the assignment.
To aggregate OC-costs on a dataset, we average the image-level OC-costs over all images.

\subsection{Optimal Transportation Problem}
We formulate the problem of finding assignment of detections to ground truths for correction cost assessment as an optimal transportation problem.
The goal of an optimal transportation problem is to find an optimal transportation plan to move goods from a collection of suppliers to a collection of demanders.
Suppose there are $m$ suppliers and $n$ demanders.
The supplier $i$ holds $s_i$ units of goods, and the demander $j$ needs $d_j$ units of goods.
Transporting a unit of goods from a supplier $i$ to a demander $j$ costs $c_{i,j}$ which constructs a pair-wise cost matrix.
The goal is to transport all goods at the minimal total cost.
Precisely, we consider the following optimization problem to find the optimal transportation plan $\pi^*$:

\begin{align}
\pi^* = \argmin_{\pi} & \sum_{i=1}^{m} \sum_{j=1}^{n} c_{i,j} \pi_{i,j}, \\
\text{s.t.} &\sum_{i=1}^m\pi_{i,j} = d_{j}, \sum_{j=1}^n\pi_{i,j} = s_i, \\
&\sum_{i=1}^{m} s_i = \sum_{j=1}^n d_j, \\
&\pi_{i,j} \ge 0\ (i=1,\ldots,m,\ j=1,\ldots,n).
\end{align}

The exact solution can be obtained with linear programming.
In our experiments, we use an off-the-shelf solver\footnote{https://PythonOT.github.io/}.

\subsection{Unit Correction Cost}
We construct a pair-wise cost matrix each of which element represents a cost to correct a detection to a certain ground truth.
We call this a unit correction cost.
Let $b_i$ be the $i$-th bounding box and $l_i$ be the bounding box's category label.
$p_i$ is a confidence score for the $i$-th detection.
A unit cost to correct one detection $i$ to a ground truth $j$ is a weighted sum of localization and classification costs as:
\begin{equation}
c_{i,j} = \lambda c_\mathrm{loc}(b_i, b_j) + (1- \lambda) c_\mathrm{cls}(p_i, l_i, l_j),
\end{equation}
where $\lambda \in [0,1]$ is a hyper-parameter to balance localization and classification costs.
If the downstream task is localization error sensitive, \eg, autonomous driving, larger $\lambda$ is recommended.

We define a unit localization cost $c_\mathrm{loc}(b_i, b_j) \in [0,1]$ as
\begin{equation}
c_\mathrm{loc}(b_i, b_j) = \frac{1-\mathrm{GIoU}(b_i, b_j)}{2},%
\end{equation}
where $\mathrm{GIoU}$ is the generalized IoU \cite{Rezatofighi_2018_CVPR}.
When two bounding boxes are identical, the unit localization cost is zero.
Without IoU thresholding, our proposed measure smoothly changes along with the localization quality.

A classification cost is defined as
\begin{equation}
c_\mathrm{cls}(p_i, l_i, l_j)=
\begin{cases}
  \frac{1 - p_i}{2} , & \text{if}\ l_i=l_j, \\
  \frac{1 + p_i}{2} , & \text{otherwise}.
\end{cases}
\end{equation}
For correctly labeled detections, higher confidence scores are rewarded more, while misclassification is heavily penalized.
Note that the unit cost is the sum of localization and classification costs, we evaluate the localization quality of misclassified detections.

\begin{figure}[t!]
\begin{minipage}{.48\linewidth}
\centering
$\beta=0.3$
\includegraphics[width=\linewidth]{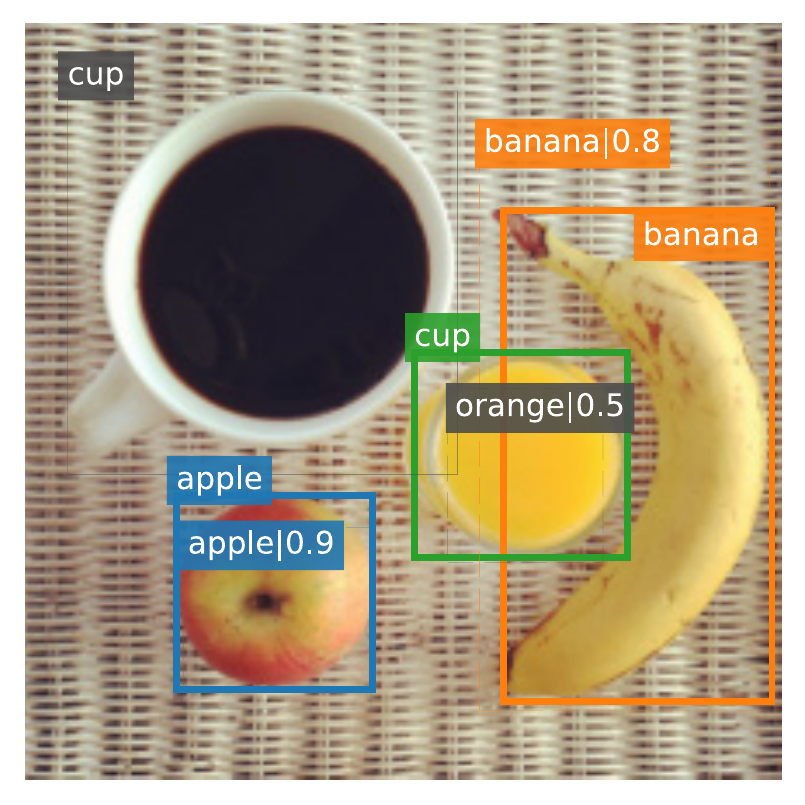}
\end{minipage}
\begin{minipage}{.48\linewidth}
\centering
$\beta=0.6$
\includegraphics[width=\linewidth]{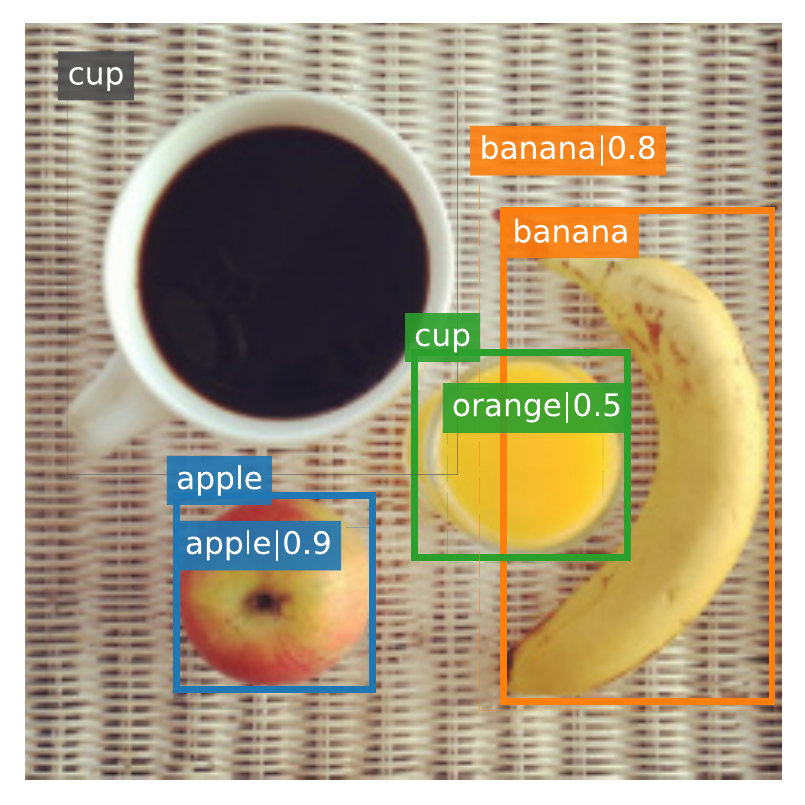}
\end{minipage}
\caption{
Assignment results with $\beta=0.3$ and $\beta=0.6$.
A ground truth (solid line) and a detection (dashed line) with the same color are associated with each other.
Gray represent that the ground truth or detections are not associated with any detections or ground truths.
The parameter $\beta$ controls the upper limit of the cost that a matched pair can take.
Left: With smaller $\beta$, the detection incorrectly labeled as orange is considered as a false positive.
Right: Larger $\beta$ allow a detection with some errors to be associated with a ground truth.
}
\label{fig:alignment}
\end{figure}
There are two types of erroneous detections.
A false positive detection refers to incorrectly detecting an instance, \eg, due to detection in background, misclassification or poor localization.
Following Pascal VOC and COCO, when there are multiple detections of the same instance, we consider the redundant detections except the best one as false positives.
A false negative refers to the error of not detecting a ground truth instance.
While finding the optimal assignment, false positive and negative detections have to be properly handled because involving false positives or negatives in the assignment can break good matches of detections and ground truths.
To this end, we introduce a dummy detection and a dummy ground truth.
False positive detections are represented by a transportation from a detection to the dummy ground truth, and false negatives are represented by transportation from the dummy detection to a ground truth.
The unit cost to transport to or from the dummy is a parameter $\beta$.
The parameter $\beta$ controls the level of an acceptable error for each assignment.
With a smaller $\beta$, a detection is more likely to be assigned to the dummy because assigning a poor detection to a ground truth costs more than assigning it to a dummy.%
With a larger $\beta$, the requirement for an assignment is relaxed to some extent.%
When a downstream task requires precise detections, using a smaller $\beta$ is recommended.

\Cref{fig:alignment} shows assignment results with different $\beta$.
Associated detections and ground truths are presented with the same color.
Gray indicates that the detection or the ground truth is assigned with the dummy.
As in \cref{fig:alignment}, with a smaller $\beta$, the detection with a classification error, which is labeled as orange, is assigned with the dummy, thus the detection is considered as a false positive.
On the other hand, with a larger $\beta$, the classification error is allowed to some extent, and the detection is associated with a ground truth.
We keep the parameters $\lambda$ and $\beta$ controllable so that developers can tune OC-cost for their problems.

\subsection{Correction Cost Computation}
Suppose a detector produces $m$ detections and the image holds $n$ ground truths.
In our formulation, detections are represented by suppliers, and ground truths are represented by demanders.
With the dummy detection, we have $m+1$ suppliers that hold $s_i$ units where $i=1, \ldots, m+1$.
In the same way, the demanders need $d_j$ units where $j=1, \ldots, n+1$.
We set the capacity of real suppliers ($s_1,\ldots,s_m$) and demanders ($d_1,\ldots,d_n$) to 1.
The capacity of the dummy supplier $s_{m+1}$ is set to $n$ and the capacity of the dummy demander $d_{n+1}$ is set to $m$.
After computing the correction cost matrix, we obtain the optimal assignment, \ie, transportation plan $\pi^*$, by solving an optimal transportation problem.
Based on the optimal assignment, we compute the correction cost.
In this computation, the cost of dummy to dummy transportation is ignored, thus we set $\pi_{m+1,n+1}$ to 0.
The transportation plan is normalized based on the remaining values.
\begin{equation}
    \tilde{\pi}_{i,j} = \frac{\pi^*_{i,j}}{\sum_{i=1}^{m} \sum_{j=1}^{n} \pi^*_{i,j}}.
\end{equation}
The final OC-cost $\Tilde{c}$ is computed as:
\begin{equation}
    \tilde{c} = \sum_{i=1}^{m+1} \sum_{j=1}^{n+1} c_{i,j} \tilde{\pi}_{i,j}.
\end{equation}

Although the computational cost increases as the number of detections per image increases, the processing time remains reasonable for a practical number of detections.
When the number of detections is equal to the number of ground truths, it takes about 22 seconds to compute OC-costs on the MS COCO validation set.

\section{Experimental Results}
Experiments are done on MS COCO validation 2017 split, which has 5000 images with annotated instances of 80 categories \cite{10.1007/978-3-319-10602-1_48}.
We test five off-the-shelf detectors: Faster-RCNN \cite{NIPS2015_5638}, RetinaNet \cite{Lin_2017_ICCV}, DETR \cite{10.1007/978-3-030-58452-8_13}, YOLOF \cite{Chen_2021_CVPR}, and VFNet \cite{Zhang_2021_CVPR}.
All the detectors use the ResNet-50 backbone.
We use pretained weights provided by MMDetection \cite{mmdetection}.

\subsection{Toy Examples}
\label{sec:toy_examples}
\begin{figure}[t!]
\begin{center}
\includegraphics[width=\linewidth]{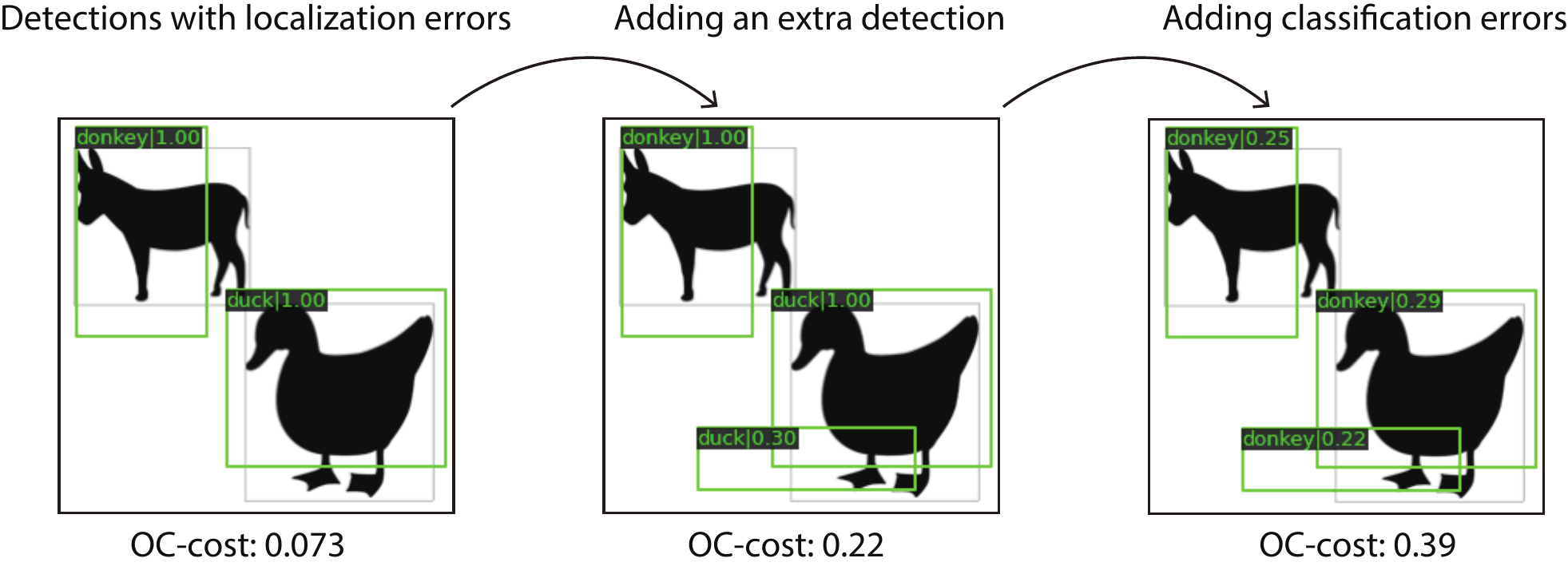}
\end{center}
\caption{
Different types of errors are sequentially added to detections. OC-cost monotonically increases at every step.
}
\label{fig:toy_example}
\end{figure}
To demonstrate how various types of errors affect OC-cost, we compute OC-cost on toy examples.
\Cref{fig:toy_example} shows three detection examples.
From left to right, we sequentially add different types of detection errors to the ground truths.
The left one shows the detections with small localization errors.
The OC-cost remains low for the detections.
In addition to the localization error, the middle one gets a false positive detection that increases the OC-cost by 0.147.
In mAP, the effect of this false positive is determined depending on detections in other images, and thus this error may or may not affect mAP.
In the right example, we perturb the predicted labels and confidence scores.
This error also boosts the OC-cost.
As these examples shows, OC-cost smoothly changes along with various types of image-level detection errors.

\subsection{Agreement with Human Preference}
\label{sec:agreement_w_human}
\begin{figure}[t!]
\begin{center}
\includegraphics[width=\linewidth]{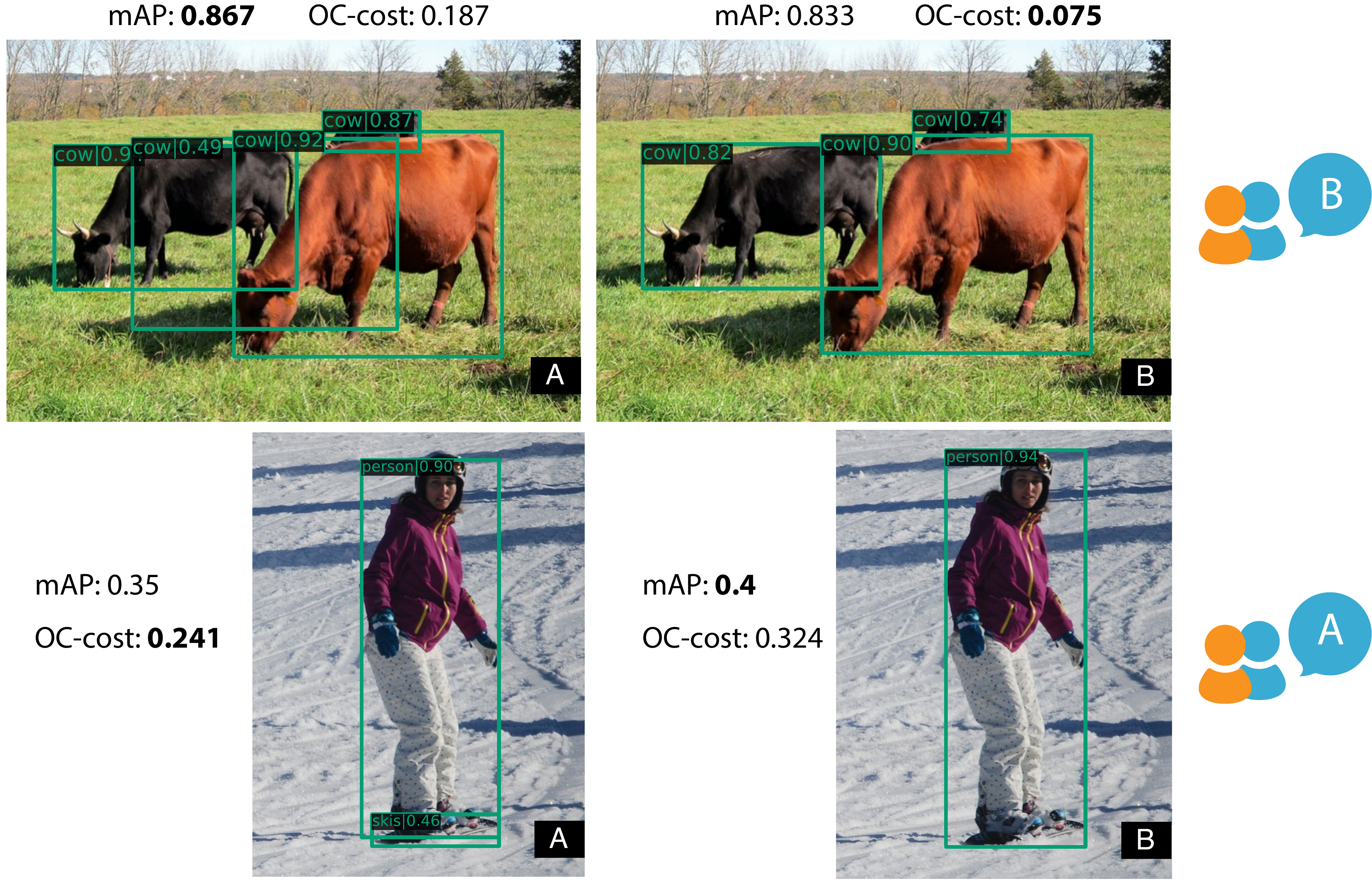}
\end{center}
\caption{
Examples of annotators' votes.
Annotators compare two detection results and vote which one looks better.
For each detection results, mAP for single image and OC-cost are displayed.
OC-cost shows better agreement with human preference.
}
\label{fig:userstudy_example}
\end{figure}
\begin{table}[t!]
\centering
\begin{tabular}{@{}lcccc@{}}
\toprule
           &                                                              & OC-cost                                                      &                                                              & mAP   \\
($\lambda$, $\beta$) & (0.2, 0.6) & (0.5, 0.6) & (0.5, 0.3) &       \\ \midrule
Accuracy   & 0.795                                                        & \textbf{0.806}                                                        & 0.58                                                         & 0.696 \\ \bottomrule
\end{tabular}
\caption{
Agreement with human preference. We compute OC-cost and mAP for a single image and compare the measures' preference to human's. The best accuracy is achieved by OC-cost at $\lambda=0.5$ and $\beta=0.6$.
} %
\label{tab:agreement_w_human}
\end{table}
\begin{figure*}[t!]
\begin{center}
\includegraphics[width=.8\linewidth]{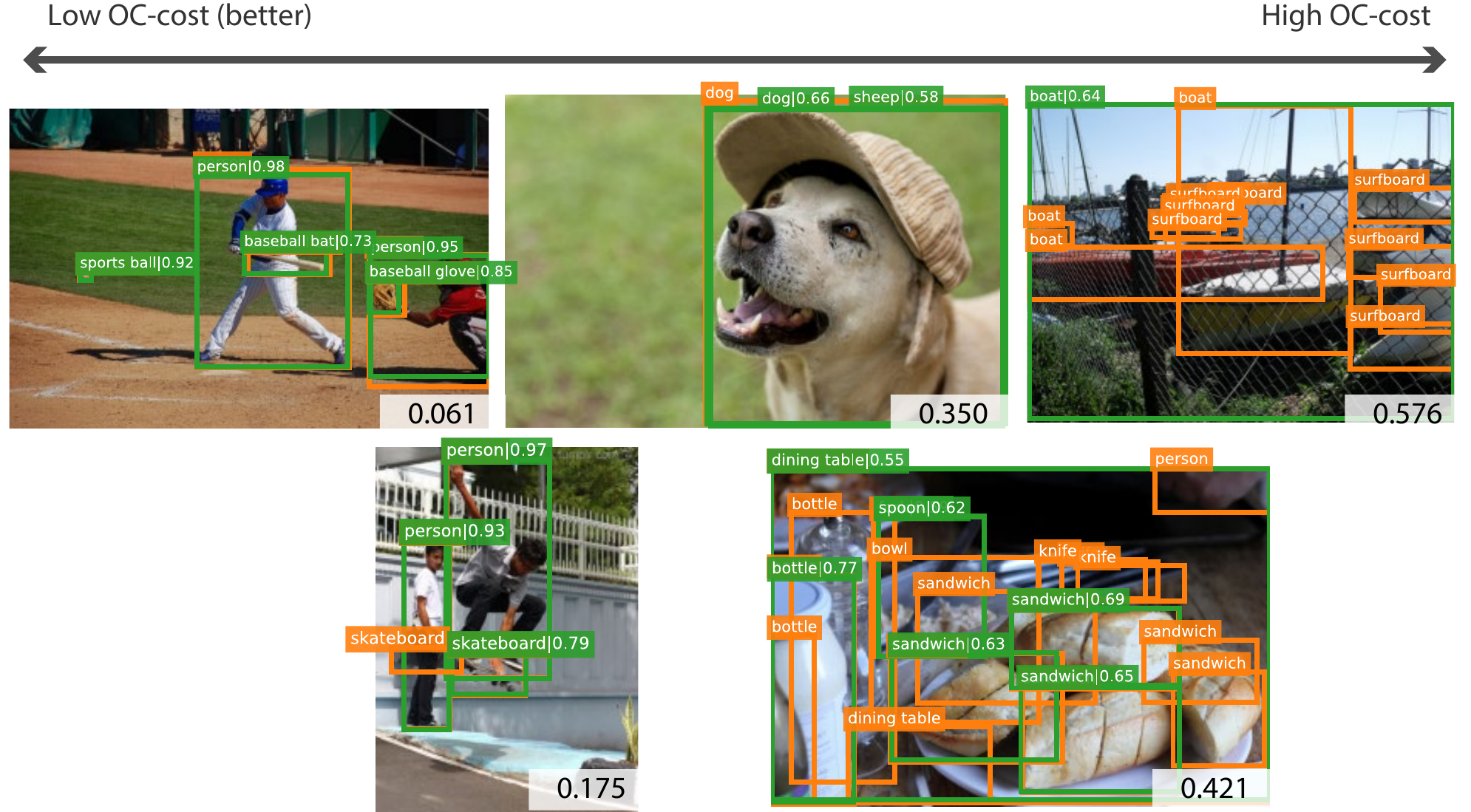}
\end{center}
\caption{
OC-cost examples. The parameters $\lambda$ is 0.5, and $\beta$ is 0.6. The detections (green) are produced by VFNet, and NMS is tuned on OC-cost. Ground truths are represented by orange bounding boxes.
OC-cost is displayed on the right bottom of each image.
}
\label{fig:realworld_example}
\end{figure*}
We validate OC-cost's agreement with human preference.
Annotators compare the results of two detectors and vote for the result which looks better.
For each compared detections, we calculate OC-costs and check if the measure %
has agreement with the human annotators.
We use RetinaNet and YOLOF as detectors, and sample 1057 images from the validation set.
We choose the samples for which OC-cost with different hyper-parameters show different preferences.
The aim of this sampling is to choose images where the two detectors differ to some extent, but the preference may not be too obvious.
Three annotators worked for the data collection.
We selected two out of three annotators with the highest Krippendorff's $\alpha$ \cite{klaus1980content}, which represents the degree of agreement.
Krippendorff's $\alpha$ for the selected annotators is 0.46.
We omit samples where the votes are split and obtain 772 samples.
Figure~\ref{fig:userstudy_example} shows examples of the detection results, corresponding quantitative measures, and the annotators' votes.

We test different hyper-parameters $\lambda$ and $\beta$ for OC-cost by grid search.
As a measure of agreement, accuracy is computed as a proportion of pairs which the automatic measure and human annotators select the same detection results.
The results with three sets of parameters are shown in Table~\ref{tab:agreement_w_human}.
The highest accuracy 0.80 is obtained at $\lambda=0.5$ and $\beta=0.6$.
We also evaluate mAP for a single image.
In a similar way to the standard mAP, we can obtain category-wise precision/recall curves for detections for a single image, and compute mAP.
The accuracy of mAP for a single image is 0.69.
This suggests that OC-cost is more consistent with human preference.
In the following experiments, we use the best parameters $\lambda=0.5$ and $\beta=0.6$

\subsection{Real-world Examples}
To demonstrate OC-cost's behavior, we show detection examples on MS-COCO dataset and corresponding OC-costs in \cref{fig:realworld_example}.
From left to right, the examples are displayed in the order of OC-cost.
The top left detection example, which successfully localizes and label the instances, obtains a fairly low cost.
The top middle example correctly detects a dog but also has a redundant misclassified detection for which OC-cost penalizes.
The top and bottom right examples got high OC-costs as they failed to detect many annotated instances. %

\subsection{Evaluating Detectors}
\begin{table}[t!]
\centering
\begin{tabular}{@{}lcc@{}}
\toprule
                     & mAP ($\uparrow$)   & OC-cost ($\downarrow$) \\ \midrule
Faster-RCNN \cite{NIPS2015_5638} & 0.38          & 0.45             \\
RetinaNet \cite{Lin_2017_ICCV}   & 0.32          & 0.28             \\
DETR \cite{10.1007/978-3-030-58452-8_13}        & \textbf{0.40} & 0.57             \\
YOLOF \cite{Chen_2021_CVPR}       & 0.32          & 0.30             \\
VFNet \cite{Zhang_2021_CVPR}       & 0.37          & \textbf{0.26}    \\ \bottomrule
\end{tabular}
\caption{
mAP and OC-cost of the off-the-shelf detectors. %
mAP and OC-cost result in opposite detector's rankings.
The result indicates that mAP and OC-cost evaluates different aspects of detectors.
} %
\label{tab:off-the-shelf_detectors}
\end{table}
\begin{table}[t!]
\centering
\begin{tabular}{@{}lcc@{}}
\toprule
                     & mAP ($\uparrow$)   & OC-Cost ($\downarrow$) \\ \midrule
Faseter-RCNN \cite{NIPS2015_5638} & 0.38 & \textbf{0.45}    \\
RetinaNet \cite{Lin_2017_ICCV}   & 0.37 & 0.52    \\
DETR \cite{10.1007/978-3-030-58452-8_13}        & 0.40 & 0.57    \\
YOLOF \cite{Chen_2021_CVPR}       & 0.38 & 0.54    \\
VFNet \cite{Zhang_2021_CVPR}       & \textbf{0.44} & 0.54    \\ \bottomrule
\end{tabular}
\caption{
Performance of the off-the-shelf detectors with NMS parameters tuned on mAP.
The best detector is highlighted in bold.
} %
\label{tab:off-the-shelf_detectors_mAP}
\end{table}
\begin{figure}[t!]
\begin{center}
\includegraphics[width=\linewidth]{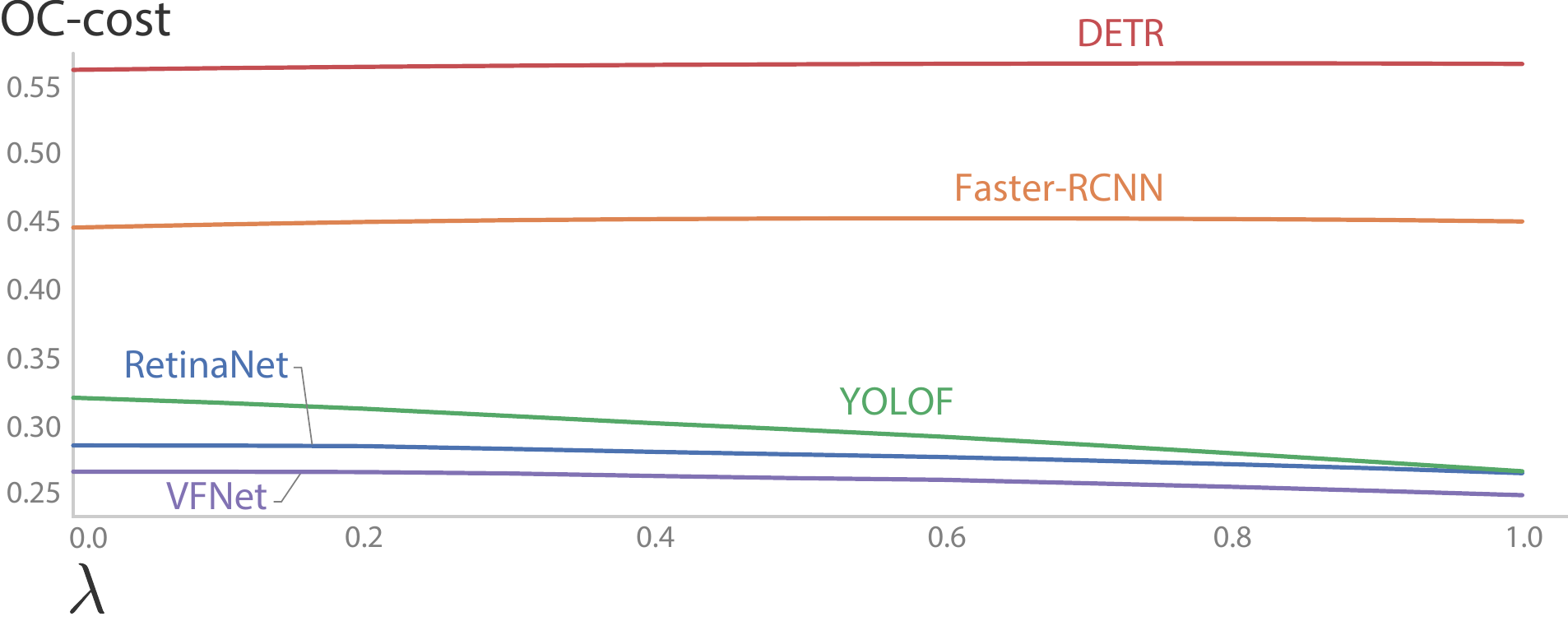}
\end{center}
\caption{
Detectors' OC-costs with different $\lambda$.
With small $\lambda$, OC-cost emphasizes classification error, while larger $\lambda$ emphasizes localization error.
OC-costs of RetinaNet, YOLOF, and VFNet decrease with larger $\lambda$.
This indicates that the error of the three detectors mainly come from classification errors
}
\label{fig:ranking_w_lambdas}
\end{figure}

We evaluate the off-the-shelf detectors in terms of mAP and OC-cost.
Each detector's hyper-parameters for Non-maximum Suppression (NMS) are tuned on OC-cost.
We does not conduct hyper-parameter tuning for DETR because it does not use NMS.
Table~\ref{tab:off-the-shelf_detectors} shows that the detectors' rankings considerably differ on mAP and OC-cost.
Note that this disagreement does not imply that one is correct and the other is not, because the measures evaluates different aspects of detections.
mAP evaluates precision/recall curves of detected instances, whereas OC-cost evaluates the per image similarity of detections and ground truths.

OC-cost rates VFNet the best when NMS parameters are tuned on OC-cost.
VFNet also performs the best in terms of mAP when NMS is tuned on mAP as shown in \cref{tab:off-the-shelf_detectors_mAP}.
This result indicates that VFNet successfully calibrates confidence scores and proposals, and we can filter out noisy detections easily by adjusting the confidence scores and the IOU threshold.
DETR results in high OC-cost, because it outputs as many detections as possible, %
which leads to a high false positive rate.
If the false positive detections are ranked lower, they hardly affect mAP.
Different from mAP, OC-cost penalizes excessive number of detections.

\Cref{fig:ranking_w_lambdas} shows the detectors' OC-costs with different $\lambda$.
A smaller $\lambda$ emphasizes classification costs, while a larger $\lambda$ emphasizes localization costs.
We observe that three detectors, RetinaNet, YOLOF, and VFNet decrease their OC-costs as $\lambda$ gets larger.
This indicates that their detection errors are mainly due to classification failures.

\subsection{Consistency Analysis}
\label{sec:consistency_analysis}
\begin{figure}[t!]
\begin{center}
\includegraphics[width=\linewidth]{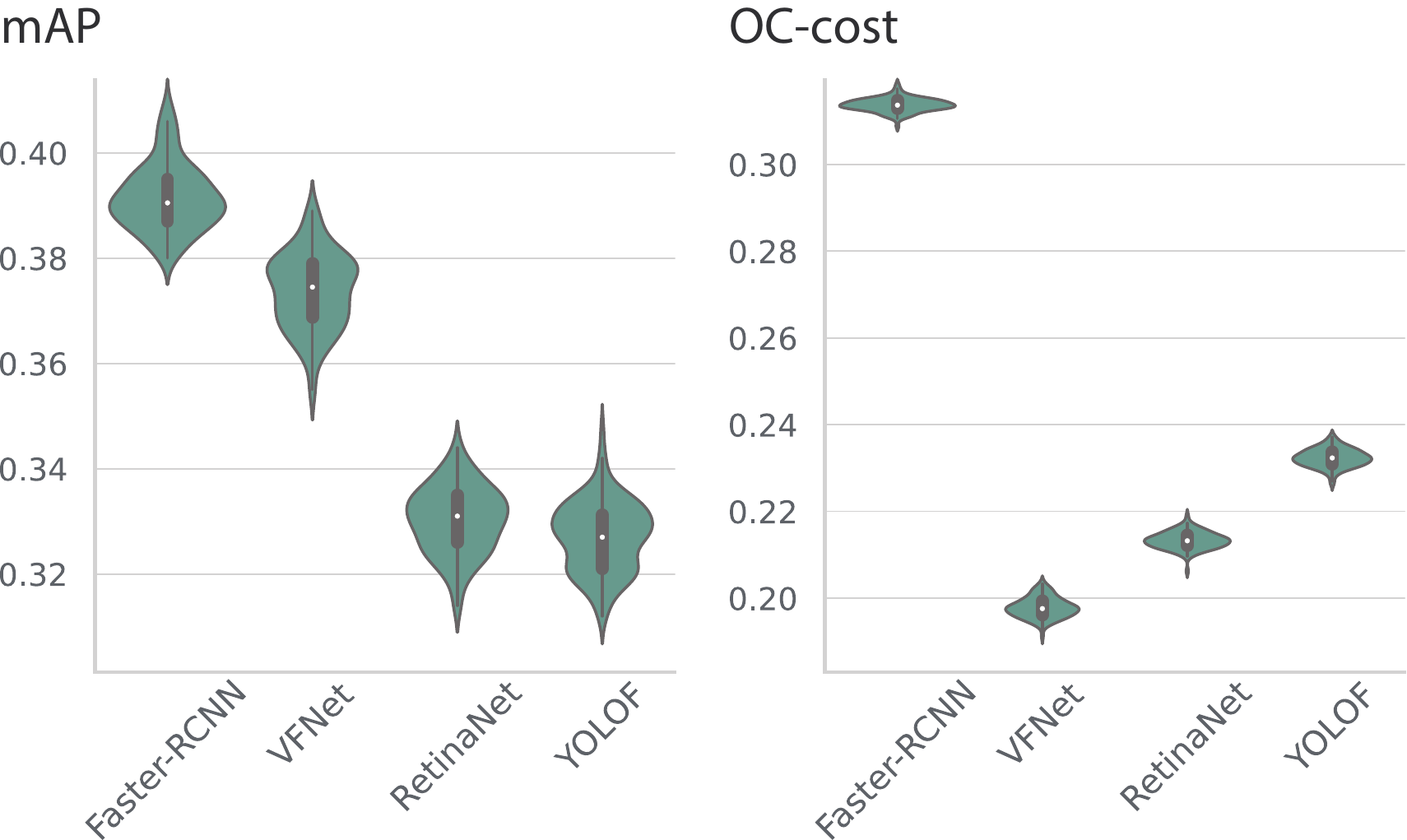}
\end{center}
\caption{
Distribution of mAP and OC-cost by bootstrapping with 100 trials.
For each trial, we randomly sample 30\% of the validation set and compute mAP and OC-cost.
The distribution's overlaps across detectors imply that the detectors' rankings likely to flip by chance.
}
\label{fig:variance}
\end{figure}
A common practice of benchmarking detectors is to test detectors on a shared dataset and compare the evaluation measure.
For reliable comparison, it is important to check that detectors rankings are not obtained by chance on a certain test set.
To investigate the consistency of detectors' rankings on different test sets, we evaluate detectors on resampled datasets and check the distribution of the evaluation measures computed on those datasets.
Specifically, for each trial, we randomly sample 30\% of MS COCO validation 2017 split with replacement and calculate the measures on the resampled data.
We repeat this process 100 times and report the distribution of sampled measurement values.

\Cref{fig:variance} shows the distributions of OC-cost and mAP.
The results of DETR is omitted for clarity.
Full results are in the supplementary material.
We observe that the distribution of mAP overlaps across detectors.
This result indicates that the performance ranking can be flipped depending on the test set.
On the other hand, OC-cost's variances remain low on smaller test sets, and the ranking of detectors is stable.
This result suggests that OC-cost enables more reliable comparison of detectors.

\subsection{Tuning NMS on OC-cost}
As \cref{tab:off-the-shelf_detectors} shows, the preferences of mAP and OC-cost are very different.
This suggests that we obtain completely different detectors depending on which measure is used to tune the detector.
To clarify this, we compare two VFNet detectors whose NMS is tuned with mAP and OC-cost.

\begin{figure}[t!]
\begin{center}
\includegraphics[width=\linewidth]{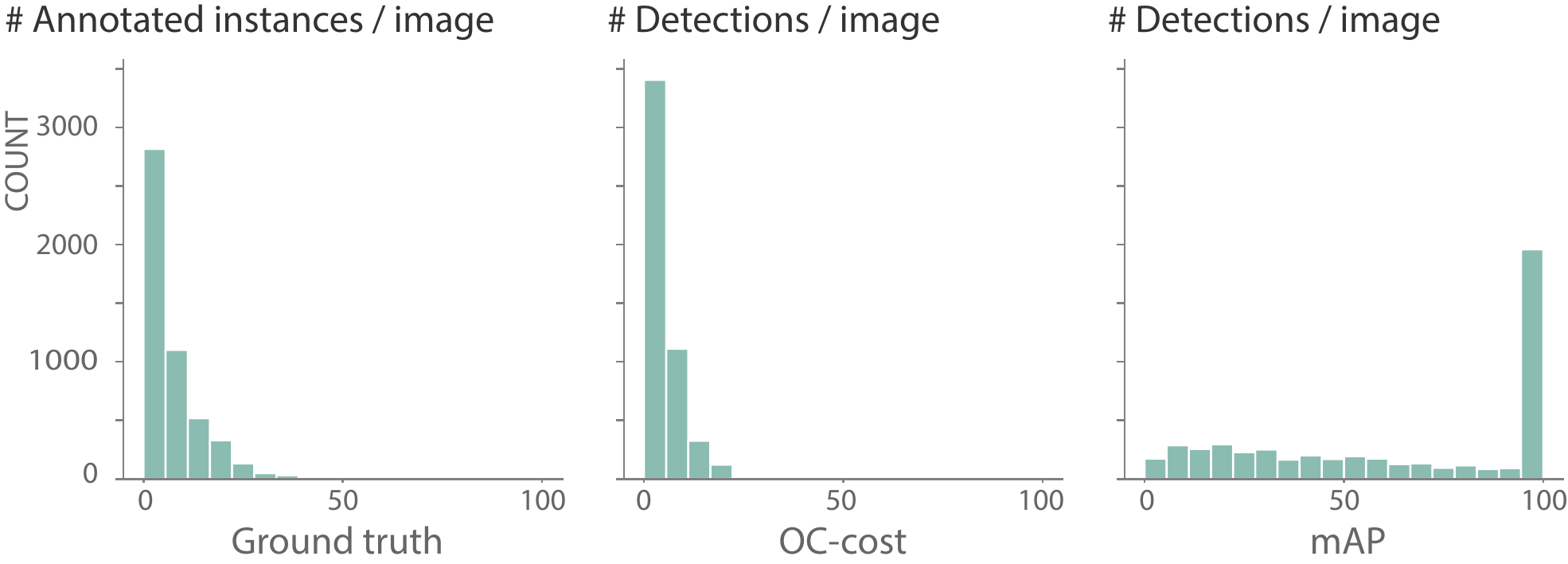}
\end{center}
\caption{
Distribution of number of detections by VFNet.
Left: The number of annotated instances. Each image has 7.2 annotated instances on average.
Center: When NMS parameters are tuned to minimize OC-cost, the number of detections are adjusted to be close to ground truths.
Right: The detector tries to outputs as many detections as possible when NMS parameters are tuned with mAP.
}
\label{fig:hptune_nbox_distribution}
\end{figure}

\begin{figure}[t]
\begin{center}
\includegraphics[width=0.95\linewidth]{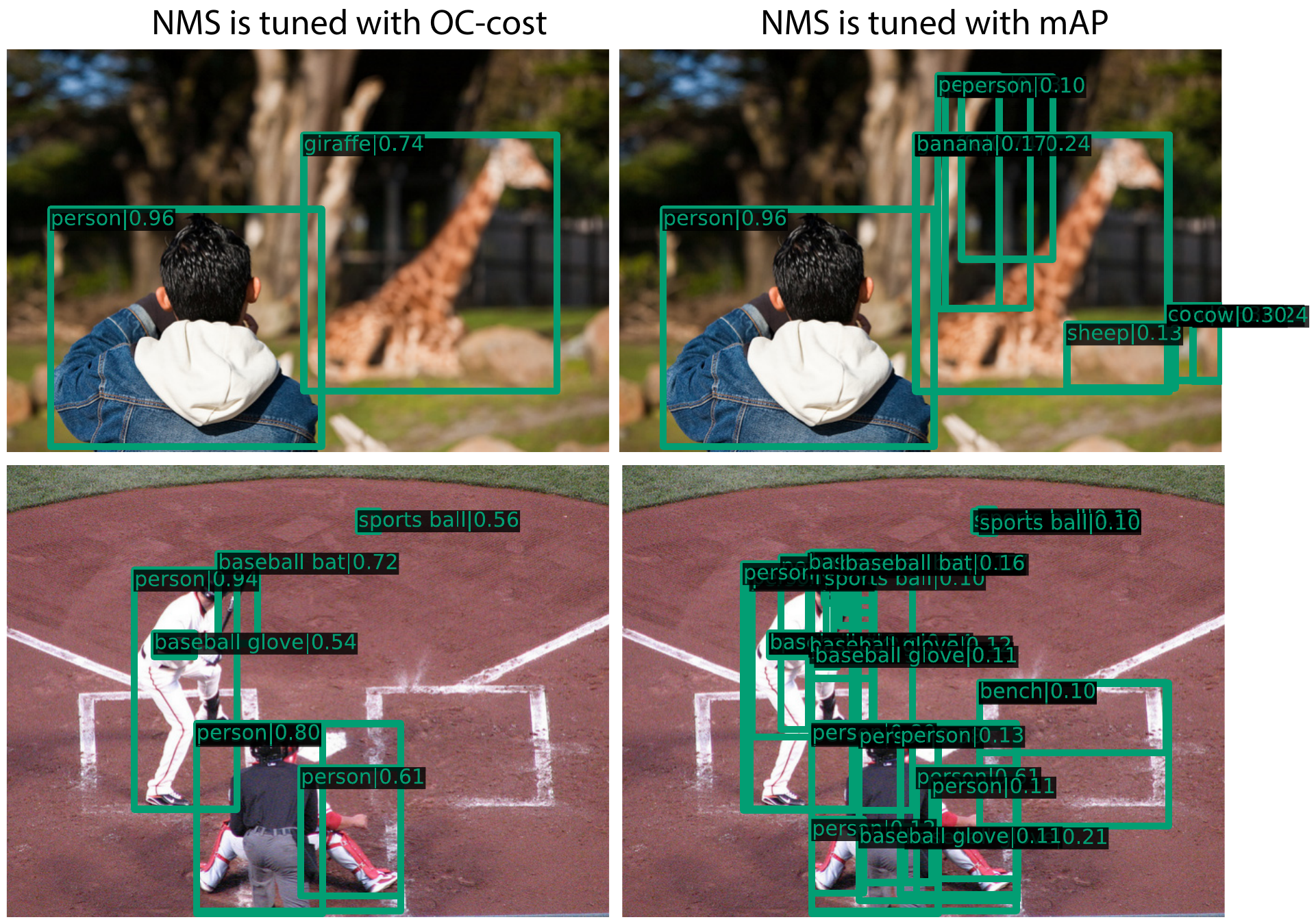}
\end{center}
\caption{
Detections with different NMS parameters.
NMS parameters are tuned with OC-cost (Left) and mAP (Right).
mAP encourages detectors to outputs many detections with low confidence scores because mAP's priority is to increase the chance to get true positives.
Unlike mAP, OC-cost penalizes false positives.
}
\label{fig:hptune_example}
\end{figure}

\Cref{fig:hptune_nbox_distribution} shows histograms of the number of detections per image.
The mAP forces the detector to make more detections in order to increase the chances of a true positive detection.
As a result, a detector tuned with mAP will make as many detections as possible in most images.
On the other hand, OC-cost penalizes false positive detections, thus the number of detections per image is adjusted to be as many as ground truths.
We can see that the distribution of the number of detections tuned with OC-cost is close to the ground truth's distribution.
\Cref{fig:hptune_example} shows examples of detections tuned on mAP and OC-cost.
OC-cost does not allow a large number of low confidence detections to be included for the purpose of increasing the recall.

mAP assumes that controlling the balance between precision and recall is each developer's responsibility. 
Thus, mAP may be useful when there is little knowledge about downstream tasks.
We recommend to tune NMS over OC-cost instead of mAP when a detector needs to avoid redundant detections.
OC-cost is particularly useful for applications that require detections to be filtered beforehand.
\section{Discussions}
\label{sec:discussions}

\paragraph{Limitations}
The main challenge in evaluation measures for object detection tasks is noise in annotations \cite{Hall_2020_WACV}.
Object detection datasets inevitably contain noise in annotations.
The noise can be introduced mainly due to ambiguity in categories/locations and annotators' skills.
Like most evaluation measures, OC-cost also assumes that ground truths are correct.
When ground-truth annotations have noise, OC-cost can underrate or overrate the detector's performance.
Designing an evaluation measure considering a noise model representing how noise is generated will be an essential topic.
Another issue is the category similarity for classification cost.
Current OC-cost treats all categories equally.
However, some applications have category-dependent risk of misclassification, \eg, misclassifying a car into a truck is less critical than a person into a traffic light in autonomous driving.
Considering category relationships is also an important issue of LVIS benchmark \cite{Gupta_2019_CVPR}.

\paragraph{Social impact}
Many real-world applications use object recognition technologies, and users of such applications are found in various regions, with different economic situations and cultural backgrounds.
Object recognition technologies have to work well for all users, and if a technology disregards any group of users, such inequality has to be detected.
A prior work reported that popular cloud services drop their performance for images from certain groups of users \cite{Vries_2019_CVPR_Workshops}.
Per-image evaluation like OC-cost is useful to detect performance degradation in those cases.

The choice of the evaluation measure has a significant impact on the detector's behavior.
This choice should be based on the various conditions of the final application.
However, due to the complexity of such decision, it is possible to make a mistake in selection of a evaluation measure or hyper-parameters.
Misuse of the evaluation measure can increase the risk of failure in the final application.
To mitigate these problems, providing guidelines and typical application scenarios would be useful so that users understand the effects and limitations of using the measure intuitively.
\section{Conclusions}
We introduce a novel measure, OC-cost, for evaluating object detectors.
We define a cost to correct detections to ground truths as a performance measure.
The experimental results demonstrate that OC-cost is consistent with human preference to some extent.
We also demonstrate that OC-cost has the capability to facilitate a fair comparison.

As we discussed, OC-cost and mAP evaluate detectors based on different assumptions.
mAP focuses on evaluating ranking performance over a dataset, while OC-cost evaluates the per-image detection accuracy.
For deeper understanding of detectors, we recommend using multiple evaluation measures that have different evaluation policies.

OC-cost can be considered as a dissimilarity measure for labeled bounding boxes.
Not only for detector evaluation, extending OC-cost for other applications is an interesting future direction.
A potential application is layout generation.
OC-cost-like measures would be helpful in sampling or retrieving layouts \cite{Patil_2021_CVPR} and analysis of generated layouts.

\paragraph{Acknowledgement}
This work was partly supported by Academy of Finland project No.~324346, JST CREST Grant No.~JPMJCR20D3 and FOREST Grant No.~JPMJFR216O, Japan.

{
    \clearpage
    \small
    \bibliographystyle{ieee_fullname}
    \bibliography{macros,main}

\begin{thebibliography}{10}\itemsep=-1pt

\bibitem{10.1145/2070781.2024192}
Nicolas Bonneel, Michiel van~de Panne, Sylvain Paris, and Wolfgang Heidrich.
\newblock Displacement interpolation using lagrangian mass transport.
\newblock {\em ACM Transactions on Graphics}, 30(6):1–12, 2011.

\bibitem{10.1007/978-3-030-58452-8_13}
Nicolas Carion, Francisco Massa, Gabriel Synnaeve, Nicolas Usunier, Alexander
  Kirillov, and Sergey Zagoruyko.
\newblock End-to-end object detection with transformers.
\newblock In {\em Proceedings of the European Conference on Computer Vision
  (ECCV)}, pages 213--229, 2020.

\bibitem{mmdetection}
Kai Chen, Jiaqi Wang, Jiangmiao Pang, Yuhang Cao, Yu Xiong, Xiaoxiao Li,
  Shuyang Sun, Wansen Feng, Ziwei Liu, Jiarui Xu, Zheng Zhang, Dazhi Cheng,
  Chenchen Zhu, Tianheng Cheng, Qijie Zhao, Buyu Li, Xin Lu, Rui Zhu, Yue Wu,
  Jifeng Dai, Jingdong Wang, Jianping Shi, Wanli Ouyang, Chen~Change Loy, and
  Dahua Lin.
\newblock {MMDetection}: Open mmlab detection toolbox and benchmark.
\newblock {\em arXiv preprint arXiv:1906.07155}, 2019.

\bibitem{Chen_2021_CVPR}
Qiang Chen, Yingming Wang, Tong Yang, Xiangyu Zhang, Jian Cheng, and Jian Sun.
\newblock You only look one-level feature.
\newblock In {\em Proceedings of the IEEE/CVF Conference on Computer Vision and
  Pattern Recognition (CVPR)}, pages 13039--13048, 2021.

\bibitem{Dave2021}
Achal Dave, Piotr Doll{\'{a}}r, Deva Ramanan, Alexander Kirillov, and Ross
  Girshick.
\newblock {Evaluating} large-vocabulary object detectors: {The} devil is in the
  details.
\newblock {\em arXiv preprint arXiv:2102.01066}, 2021.

\bibitem{Vries_2019_CVPR_Workshops}
Terrance de Vries, Ishan Misra, Changhan Wang, and Laurens van~der Maaten.
\newblock Does object recognition work for everyone?
\newblock In {\em Proceedings of the IEEE/CVF Conference on Computer Vision and
  Pattern Recognition (CVPR) Workshops}, 2019.

\bibitem{Everingham2010}
Mark Everingham, Luc {Van Gool}, Christopher K~I Williams, John Winn, and
  Andrew Zisserman.
\newblock {The Pascal Visual Object Classes (VOC) Challenge}.
\newblock {\em International Journal of Computer Vision}, 88(2):303--338, 2010.

\bibitem{Ge_2021_CVPR}
Zheng Ge, Songtao Liu, Zeming Li, Osamu Yoshie, and Jian Sun.
\newblock {OTA}: {Optimal} transport assignment for object detection.
\newblock In {\em Proceedings of the IEEE/CVF Conference on Computer Vision and
  Pattern Recognition (CVPR)}, pages 303--312, 2021.

\bibitem{Gupta_2019_CVPR}
Agrim Gupta, Piotr Dollar, and Ross Girshick.
\newblock {LVIS}: {A} dataset for large vocabulary instance segmentation.
\newblock In {\em Proceedings of the IEEE/CVF Conference on Computer Vision and
  Pattern Recognition (CVPR)}, 2019.

\bibitem{Hall_2020_WACV}
David Hall, Feras Dayoub, John Skinner, Haoyang Zhang, Dimity Miller, Peter
  Corke, Gustavo Carneiro, Anelia Angelova, and Niko Suenderhauf.
\newblock Probabilistic object detection: Definition and evaluation.
\newblock In {\em Proceedings of the IEEE/CVF Winter Conference on Applications
  of Computer Vision (WACV)}, pages 1031--1040, 2020.

\bibitem{TIDE_derek_2012}
Derek Hoiem, Yodsawalai Chodpathumwan, and Qieyun Dai.
\newblock Diagnosing error in object detectors.
\newblock In {\em Proceedings of the European Conference on Computer Vision
  (ECCV)}, pages 340--353, 2012.

\bibitem{klaus1980content}
Krippendorff Klaus.
\newblock Content analysis: An introduction to its methodology, 1980.

\bibitem{Lin_2017_ICCV}
Tsung-Yi Lin, Priya Goyal, Ross Girshick, Kaiming He, and Piotr Dollar.
\newblock Focal loss for dense object detection.
\newblock In {\em Proceedings of the IEEE International Conference on Computer
  Vision (ICCV)}, pages 2980--2988, 2017.

\bibitem{10.1007/978-3-319-10602-1_48}
Tsung-Yi Lin, Michael Maire, Serge Belongie, James Hays, Pietro Perona, Deva
  Ramanan, Piotr Doll{\'a}r, and C.~Lawrence Zitnick.
\newblock {Microsoft COCO}: {Common} objects in context.
\newblock In {\em Proceedings of the European Conference on Computer Vision
  (ECCV)}, pages 740--755, 2014.

\bibitem{LRP_kemal_2018}
Kemal Oksuz, Baris~Can Cam, Emre Akbas, and Sinan Kalkan.
\newblock {Localization Recall Precision (LRP)}: {A} new performance metric for
  object detection.
\newblock In {\em Proceedings of the European Conference on Computer Vision
  (ECCV)}, pages 521--537, 2018.

\bibitem{OZDEMIR20101128}
Bahadır {\"{O}}zdemir, Selim Aksoy, Sandra Eckert, Martino Pesaresi, and
  Daniele Ehrlich.
\newblock Performance measures for object detection evaluation.
\newblock {\em Pattern Recognition Letters}, 31(10):1128--1137, 2010.

\bibitem{Patil_2021_CVPR}
Akshay~Gadi Patil, Manyi Li, Matthew Fisher, Manolis Savva, and Hao Zhang.
\newblock {LayoutGMN: Neural} graph matching for structural layout similarity.
\newblock In {\em Proceedings of the IEEE/CVF Conference on Computer Vision and
  Pattern Recognition (CVPR)}, pages 11048--11057, 2021.

\bibitem{NIPS2015_5638}
Shaoqing Ren, Kaiming He, Ross Girshick, and Jian Sun.
\newblock {Faster R-CNN}: {Towards} real-time object detection with region
  proposal networks.
\newblock In {\em Advances in Neural Information Processing Systems (Neurips)},
  pages 91--99, 2015.

\bibitem{Rezatofighi_2018_CVPR}
Hamid Rezatofighi, Nathan Tsoi, JunYoung Gwak, Amir Sadeghian, Ian Reid, and
  Silvio Savarese.
\newblock Generalized intersection over union.
\newblock In {\em The IEEE Conference on Computer Vision and Pattern
  Recognition (CVPR)}, pages 658--666, 2019.

\bibitem{voorhees2002overview}
Ellen~M Voorhees and Donna Harman.
\newblock Overview of {TREC} 2002.
\newblock In {\em Trec}, 2002.

\bibitem{Zhang_2021_CVPR}
Haoyang Zhang, Ying Wang, Feras Dayoub, and Niko Sunderhauf.
\newblock Varifocalnet: An iou-aware dense object detector.
\newblock In {\em Proceedings of the IEEE/CVF Conference on Computer Vision and
  Pattern Recognition (CVPR)}, pages 8514--8523, 2021.

\end{thebibliography}
}

\appendix

\setcounter{page}{1}

\twocolumn[
\centering
\Large
\textbf{Optimal Correction Cost for Object Detection Evaluation} \\
\vspace{0.5em}Supplementary Material \\
\vspace{1.0em}
] %
\appendix

\section{Details of the Annotation Experiment}
The three annotators in \cref{sec:agreement_w_human} are employed as our in-house annotation team. 
We explained the purpose of the project to the annotators in advance, and they were able to ask any questions during the work.
Each annotator completed annotating 1057 samples in four days.
They reviewed paired detection results for a subset of the COCO Detection dataset \cite{10.1007/978-3-319-10602-1_48} and assigned a binary preference to each pair.
We did not store any personal information for this project.
We believe that this annotation task does not violate the ethical principles in the CVPR ethics guidelines.
We do not show the annotation interface in this supplementary material because it may reveal the authors' identity.

\section{Full Results of Consistency Analysis}
\begin{figure}[h]
\begin{center}
\includegraphics[width=\linewidth]{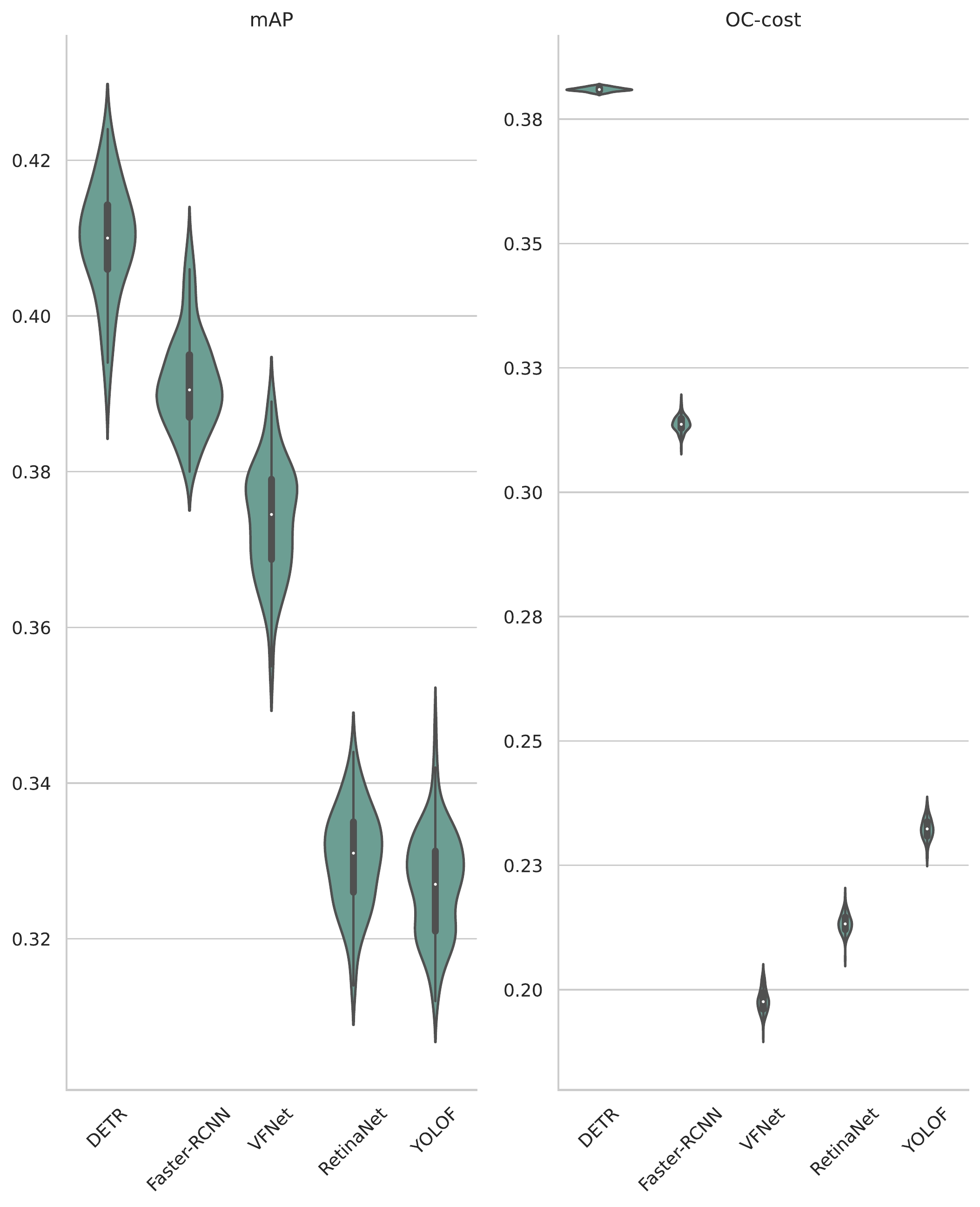}
\end{center}
\caption{
The full results of \cref{fig:variance}.
Distributions of mAP and OC-cost are obtained by bootstrapping with 100 trials.
The distributions' overlaps across detectors imply that the detectors' rankings likely to flip by chance.
}
\label{fig:variance_full}
\end{figure}
We omit the DETR's result in \cref{fig:variance} for visibility.
We show full results in \cref{fig:variance_full}.
The DETR's result does not change our conclusion that OC-cost's detectors' rankings are more stable than mAP.

\section{Interactive Demo}
\begin{figure}[h!]
\begin{center}
\includegraphics[width=\linewidth]{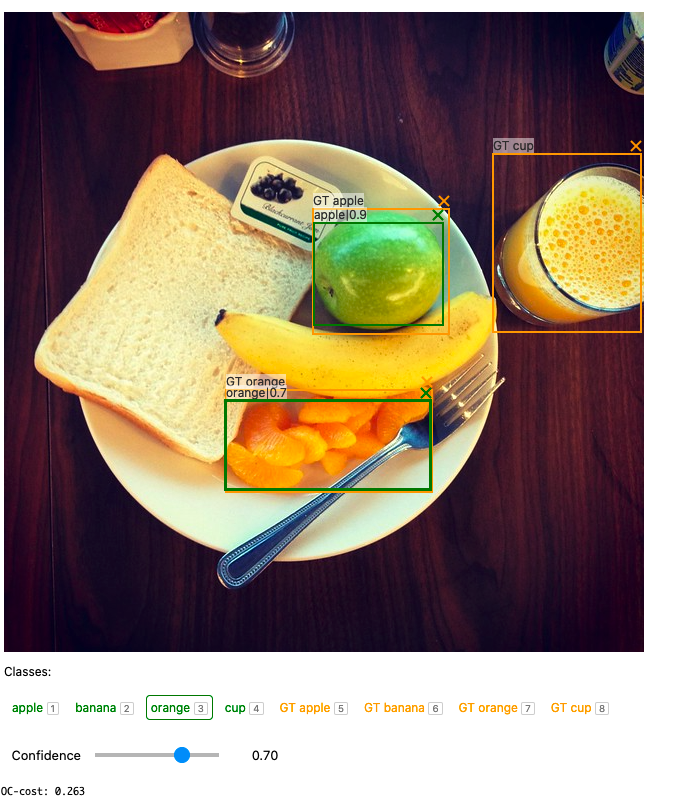}
\end{center}
\caption{
We can interactively give ground truths with orange boxes and detections with green ones.
Once the ground truths and the detections are modified, corresponding OC-cost is displayed below the image.
}
\label{fig:demo}
\end{figure}
We attach to this supplementary material a python notebook for an interactive demo.
The screenshot of the demo is in \cref{fig:demo}.
In the demo, OC-costs are computed for different detections and ground truths.

\section{OC-cost Examples}
\begin{figure*}[t!]
\begin{center}
\includegraphics[width=\textwidth,height=\textheight]{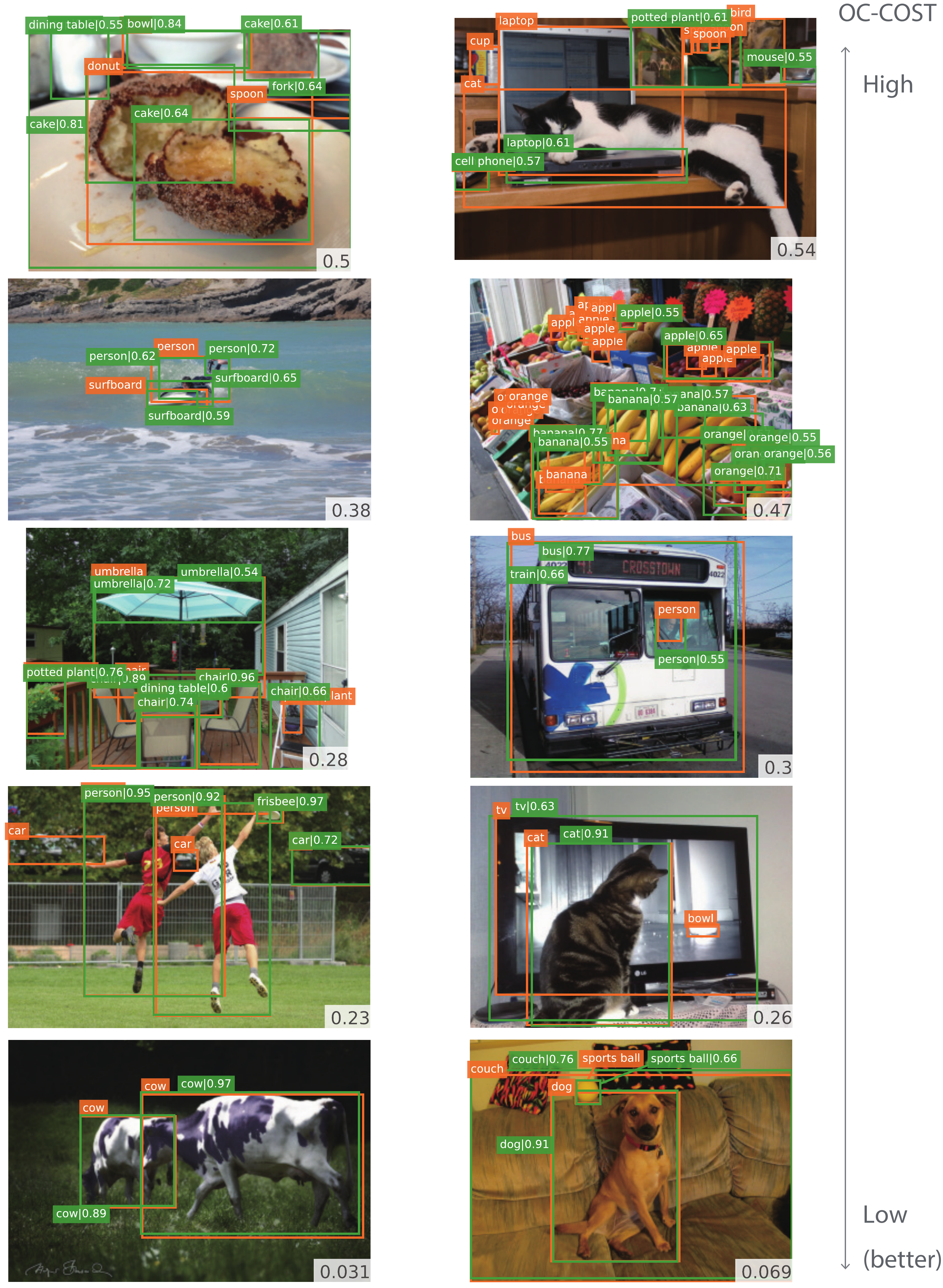}
\end{center}
\caption{
OC-cost examples. The parameters $\lambda$ is 0.5, and $\beta$ is 0.6. The detections (green) are produced by VFNet, and NMS is tuned on OC-cost. Ground truths are represented by orange bounding boxes.
OC-cost is displayed on the right bottom of each image.
}
\label{fig:spl_example}
\end{figure*}
We showcase detection examples on MS-COCO dataset and corresponding OC-costs in \cref{fig:spl_example}.
From top to bottom, the examples are displayed in the order of OC-cost.
The parameters $\lambda$ is 0.5, and $\beta$ is 0.6. The detections (green) are produced by VFNet, and NMS is tuned on OC-cost. Ground truths are represented by orange bounding boxes.

\end{document}